\documentclass{article} 
\usepackage{iclr2020_conference,times}

\usepackage{amsmath,amsfonts,bm}

\def\eqref#1{equation~\ref{#1}}

\def\1{\bm{1}}

\DeclareMathAlphabet{\mathsfit}{\encodingdefault}{\sfdefault}{m}{sl}
\SetMathAlphabet{\mathsfit}{bold}{\encodingdefault}{\sfdefault}{bx}{n}

\usepackage{color}
\usepackage{hyperref}
\usepackage{url}
\usepackage{multirow}
\usepackage{graphicx}
\usepackage{wrapfig}
\usepackage{microtype}

\title{Factorized Multimodal Transformer for Multimodal Sequential Learning}

\author{Amir Zadeh$^\dagger$, Chengfeng Mao$^\dagger$, Kelly Shi$^\dagger$, Yiwei Zhang$^\dagger$, Paul Liang$^\diamond$, \\ \textbf{Soujanya Poria$^\star$ \& Louis-Philippe Morency$^\dagger$} \\
$^\dagger$Language Technologies Institute, Machine Learning Department$^\diamond$, SCS, CMU\\
$^\star$Singapore University of Technology and Design\\
\texttt{\{abagherz,chengfem,jiaxins1,yiweizh2,pliang\}@cs.cmu.edu} \\
 \texttt{soujanya\_poria@sutd.edu.sg,morency@cs.cmu.edu}\\
}

\newcommand{\commentA}[1]{\textcolor{red}{Amir: #1}}
\newcommand{\commentAT}[1]{\textcolor{blue}{Amir to Chengfeng: #1}}

\iclrfinalcopy 
\begin{document}

\maketitle

\begin{abstract}
The complex world around us is inherently multimodal and sequential (continuous). Information is scattered across different modalities and requires multiple continuous sensors to be captured. As machine learning leaps towards better generalization to real world, multimodal sequential learning becomes a fundamental research area. Arguably,  modeling arbitrarily distributed spatio-temporal dynamics within and across modalities is the biggest challenge in this research area. In this paper, we present a new transformer model, called the Factorized Multimodal Transformer (FMT) for multimodal sequential learning. FMT inherently models the intramodal and intermodal (involving two or more modalities) dynamics within its multimodal input in a factorized manner. The proposed factorization allows for increasing the number of self-attentions to better model the multimodal phenomena at hand; without encountering difficulties during training (e.g. overfitting) even on relatively low-resource setups. All the attention mechanisms within FMT have a full time-domain receptive field which allows them to asynchronously capture long-range multimodal dynamics. In our experiments we focus on datasets that contain the three commonly studied modalities of language, vision and acoustic. We perform a wide range of experiments, spanning across 3 well-studied datasets and 21 distinct labels. FMT shows superior performance over previously proposed models, setting new state of the art in  the studied datasets.
\end{abstract}
\section{Introduction}

In many naturally occurring scenarios, our perception of the world is multimodal. For example, consider multimodal language (face-to-face communication), where modalities of language, vision and acoustic are seamlessly used together for communicative intent~\citep{kottur-etal-2019-clevr}. Such scenarios are widespread in everyday life, where continuous sensory perceptions form multimodal sequential data. Each modality within multimodal data exhibits exclusive \textit{intramodal} dynamics, and presents a unique source of information. Modalities are not fully independent of each other. Relations across two (bimodal) or more (trimodal, \dots) of them form \textit{intermodal} dynamics; often asynchronous spatio-temporal dynamics which bind modalities together~\citep{tensoremnlp17}.

Learning from multimodal sequential data has been an active, yet challenging research area within the field of machine learning~\citep{baltruvsaitis2018multimodal}. Various approaches relying on graphical models or RNNs have been proposed for multimodal sequential learning. Transformer models are a new class of neural models that rely on a carefully designed non-recurrent architecture for sequential modeling~\citep{vaswani2017attention}. Their superior performance is attributed to a self-attention mechanism, which is uniquely capable of highlighting related information across a sequence. This self-attention is a particularly appealing mechanism for multimodal sequential learning, as it can be modified into a strong neural component for finding relations between different modalities (the cornerstone of this paper). In practice, numerous such relations may simultaneously exist within multimodal data, which would require increasing the number of attention units (i.e. heads). Increasing the number of attentions in an efficient and semantically meaningful way inside a transformer model, can boost the performance in modeling multimodal sequential data.

In this paper, we present a new transformer model for multimodal sequential learning, called Factorized Multimodal Transformer (FMT). FMT is capable of modeling asynchronous intramodal and intermodal dynamics in an efficient manner, within one single transformer network. It does so by specifically accounting for  possible sets of interactions between modalities (i.e. factorizing based on combinations) in a Factorized Multimodal Self-attention (FMS) unit. We evaluate the performance of FMT on multimodal language: a challenging type of multimodal data which exhibits idiosyncratic and asynchronous spatio-temporal relations across language, vision and acoustic modalities. FMT is compared to previously proposed approaches for multimodal sequential learning over multimodal sentiment analysis (CMU-MOSI)~\citep{zadeh2016mosi}, multimodal emotion recognition (IEMOCAP)~\citep{Busso2008IEMOCAP:Interactiveemotionaldyadic}, and multimodal personality traits recognition (POM)~\citep{Park:2014:CAP:2663204.2663260}.
\section{Related Works}\label{sec:related}

The related works to studies in this paper fall into two main areas.

\subsection{Multimodal Sequential Learning}

Modeling multimodal sequential data is among the core research areas within the field of machine learning. In this area, previous work can be classified into two main categories. 

The first category of models, and arguably the simplest, are models that use early or late fusion. Early fusion uses feature concatenation of all modalities into a single modality. Subsequently, the multimodal sequential learning task is treated as a unimodal one and tackled using unimodal sequential models such as Hidden Markov Models (HMMs)~\citep{baum1966statistical}, Hidden Conditional Random Fields~(HCRFs) \citep{Quattoni:2007:HCR:1313053.1313265,morency2007latent}, and RNNs (e.g. LSTMs~\cite{hochreiter1997long}). While such models are often successful for real unimodal data (i.e. not feature concatenated multimodal data), they lack the necessary components to deal with multimodal data often causes suboptimal performance~\citep{xu2013survey}. Contrary to early fusion which concatenates modalities at input level, late fusion models have relied on learning ensembles of weak classifiers from different modalities~\citep{Snoek:2005:EVL:1101149.1101236,Vielzeuf:2017:TMF:3136755.3143011,Nojavanasghari:2016:DMF:2993148.2993176}. Hybrid methods have also been used to combine early and late fusion together~\citep{Wu_2019_CVPR,DBLP:journals/corr/abs-1812-00500,lazaridou-etal-2015-combining,DBLP:journals/corr/HuARDS17}.

The second category of models comprise of models specifically designed for multimodal data. Multimodal variations of graphical models have been proposed, including Multi-view HCRFs where the potentials of the HCRF are changed to facilitate multiple modalities \cite{song2012multi,song2013action}. Multimodal models based on LSTMs include Multi-view LSTMs~\cite{rajagopalan2016extending}, Memory Fusion Network~\citep{zadeh2018memory} with its recurrent and graph variants~\citep{liang2018multimodal,zadeh2018multimodal}, as well as Multi-attention Recurrent Networks~\citep{zadeh2018multi}. Studies have also proposed generic fusion techniques that can be used in various models including Tensor Fusion~\citep{tensoremnlp17} and its approximate variants~\citep{liang2019learning,liu2018efficient}, as well as Compact Bilinear Pooling~\citep{DBLP:journals/corr/GaoBZD15,fukui2016multimodal,DBLP:journals/corr/KimOKHZ16}. 

Many of these models, from both first and second categories, are used as baselines in this paper. 

\subsection{Transformer Model}

Transformer is a non-recurrent neural architecture designed for modeling sequential data~\citep{vaswani2017attention}. It has shown superior performance across multiple NLP tasks when compared to RNN-based or convolutional architectures~\citep{DBLP:journals/corr/abs-1810-04805,vaswani2017attention}. This superior performance of Transformer model is largely credited to a self-attention; a neural component that allows for efficiently extracting both short and long-range dependencies within its input sequence space. Transformer models have been successfully applied to various areas within machine learning including NLP and computer vision~\citep{DBLP:journals/corr/abs-1906-08237,DBLP:journals/corr/abs-1802-05751,alsentzer-etal-2019-publicly}. Extending transformer to multimodal domains, specially for structured multimodal sequences is relatively understudied; with the previous works mainly focusing on using transformer models for modality alignment using cross-modal links between single transformers for each modality~\citep{tsai2019multimodal}. 
\section{Factorized Multimodal Transformer (FMT) Model}
\label{sec:model}

\begin{wrapfigure}{R}{5.5cm}
\vspace{-35pt}
\begin{center}
\includegraphics[width=\linewidth]{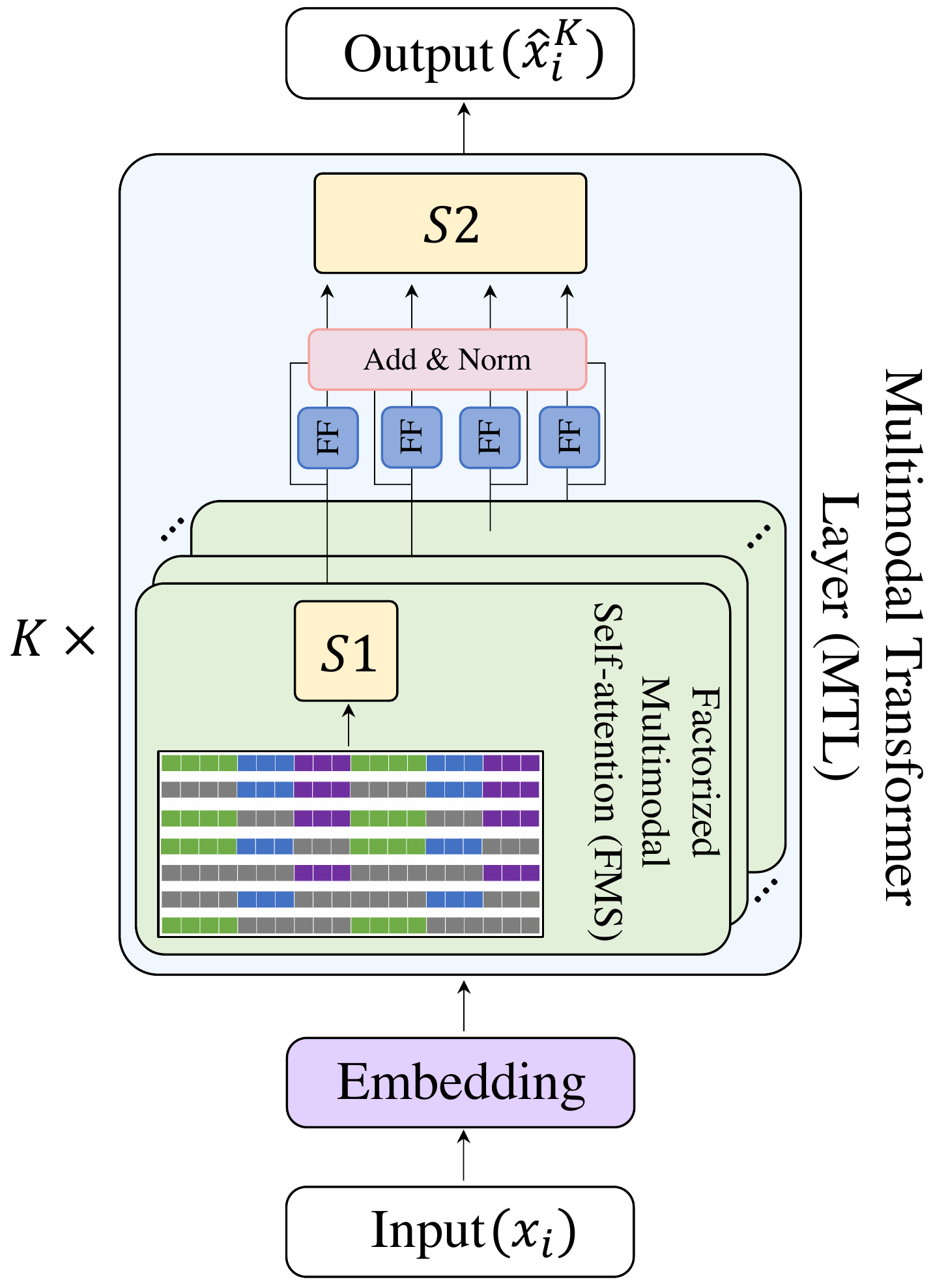}
\end{center}
\caption{Overview of the proposed Factorized Multimodal Transformer (FMT) model. }
\label{fig:fmt}
\end{wrapfigure}

In this section, we outline the proposed Factorized Multimodal Transformer\footnote{Code (release April 15th, 2020): https://github.com/A2Zadeh/Factorized-Multimodal-Transformer, Public Data: https://github.com/A2Zadeh/CMU-MultimodalSDK} (FMT). Figure \ref{fig:fmt} shows the overall structure of the FMT model. The input first goes through an embedding layer, followed by multiple Multimodal Transformer Layers (MTL). Each MTL consists of multiple Factorized Multimodal Self-attentions (FMS). FMS explicitly accounts for intramodal and intermodal factors within its multimodal input. $\mathcal{S}1$ and $\mathcal{S}2$ are two summarization networks. They are necessary components of FMT which allow for increasing the number of attentions efficiently, without overparameterization of the FMT.

\subsection{Input Embedding}

Consider a multimodal sequential dataset with constituent modalities of language, vision and acoustic. The modalities are denoted as $\{\texttt{L},\texttt{V},\texttt{A}\}$ from hereon for abbreviation. After resampling using a reference clock, modalities can follow the same frequency~\citep{chen2017multimodal}. Essentially, this resampling is often based on word timestamps (i.e. word alignment). Subsequently, the dataset can be denoted as:
\begin{align*}
    D=\Big\{x_i=\big[ x_{(t,i)} = \langle l_{(t,i)},v_{(t,i)},a_{(t,i)}\rangle \big]_{t=1}^{t=\mathcal{T}_i},y_i\Big\}_{i=1}^{N}
\end{align*}
$x_i \in \mathbb{R}^{ \mathcal{T}_i \times d_x}, y_i \in \mathbb{R}^{d_y}$ are the inputs and labels. $x_{(t,i)}=\langle l_{(t,i)},v_{(t,i)},a_{(t,i)} \rangle$ is a triplet of language, visual and audio inputs for timestamp $t$ in $i$-th datapoint. $N$ is the total number of samples within the dataset, and $\mathcal{T}_i$ the total number of timestamps within $i$-th datapoint. Zero paddings (on the left) can be used to unify the length of all sequences to a desired fixed length $\mathcal{T}$.  $d_x=d_\texttt{L}+d_\texttt{V}+d_\texttt{A}$ denotes the dimensionality of input at each timestep, which in turn is equal to the sum of dimensionality of each modality. $d_y$ denotes the dimensionality of the associated labels of a sequence. 

At the first step within the FMT model, each modality is passed to a unimodal embedding layer with the operation $\mathcal{E}_{\texttt{M} \in \{\texttt{L},\texttt{V},\texttt{A}\}}(\cdot); \mathbb{R}^{d_{\texttt{M} \in \{\texttt{L},\texttt{V},\texttt{A}\}}} \mapsto \mathbb{R}^{e_{\texttt{M} \in \{\texttt{L},\texttt{V},\texttt{A}\}}}$. In turn, $\mathcal{E}_{\texttt{M}}$ takes as input $m_{(t,i)}; m \in \{l,v,a\}$. Positional embeddings are also added to the input at this stage. The output of the embeddings collectively form $\hat{x}^0=\langle \hat{l}^0$, $\hat{v}^0$, $\hat{a}^0\rangle$. We denote the dimensionality of this output as $e_x=e_\texttt{L}+e_\texttt{V}+e_\texttt{A}$.

\subsection{Multimodal Transformer Layer (MTL)}

After the initial embedding, FMT now consists of a stack of Multimodal Transformer Layers (MTL). MTL 1) captures factorized dynamics within multimodal data in parallel, and 2) aligns the time-asynchronous information both within and across modalities. Both of these are achieved using multiple Factorized Multimodal Self-attentions (FMS), each of which has multiple specialized self-attentions inside. The high dimensionality of the intermediate attention outputs within MTL and FMS is controlled using two distinct summarization networks. The continuation of this section provides detailed explanation of the inner-operations of MTL.

\begin{figure*}[t!]
\begin{center}
\includegraphics[width=0.9\linewidth]{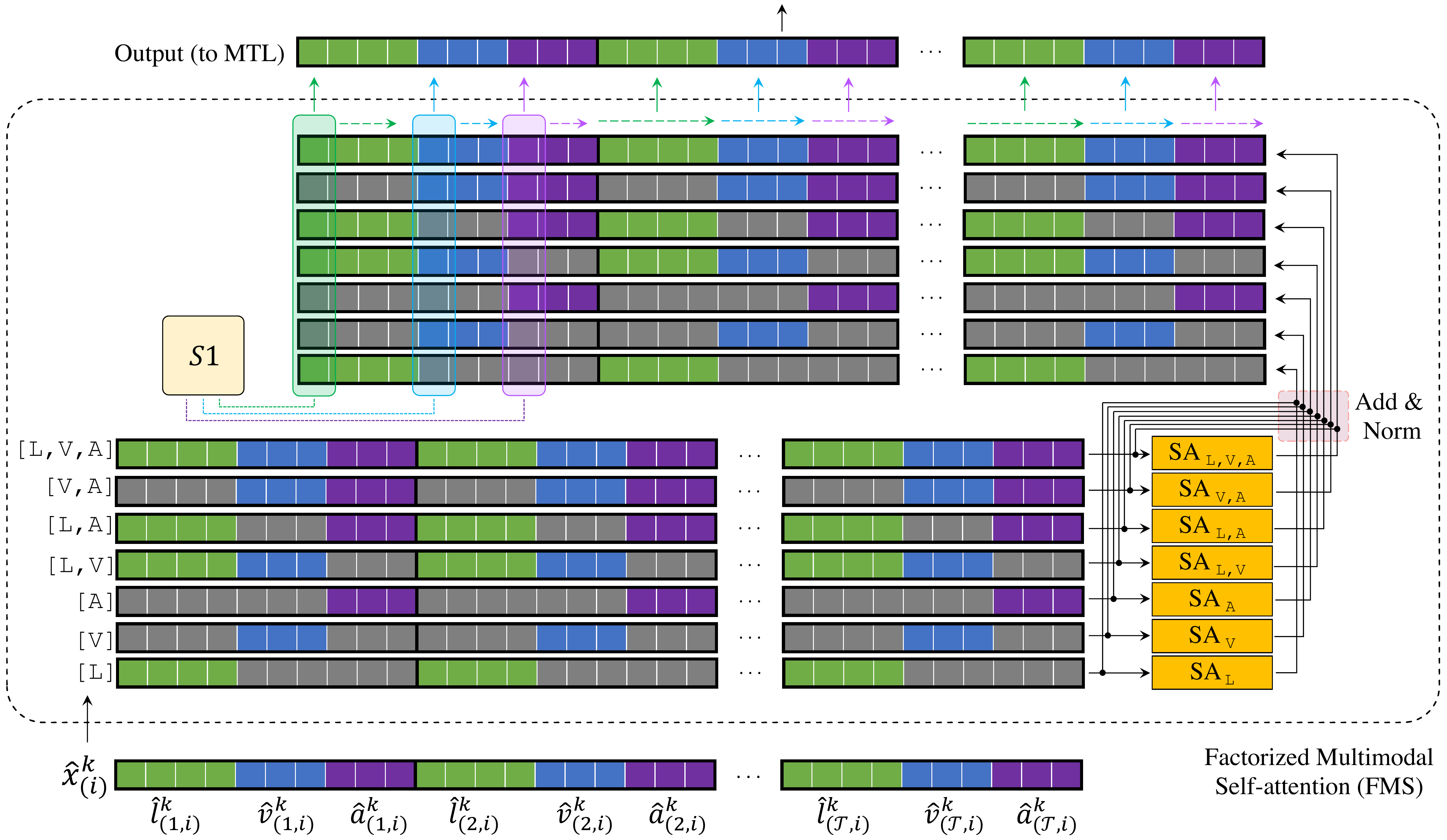}
\end{center}
\caption{Best viewed in color. Overview of a single Factorized Multimodal Self-attention (FMS) in $k$-th MTL. SA is self-attention~\citep{vaswani2017attention} with full time-domain receptive field ($0 \dots \mathcal{T}$). The grayed areas are for demonstration purposes, and not a part of the implementation. }
\label{fig:attention}
\end{figure*}

Let $\hat{x}_i^k=\langle \hat{l}^{k}_{(\cdot,i)},\hat{v}^{k}_{(\cdot,i)},\hat{a}^{k}_{(\cdot,i)}\rangle$ denote the input to the $k$-th MTL. We assume a total of $K$ MTLs in a FMT (indexed $0\dots K-1$), with $k=0$ being the output of the embedding layer (input to $k=0$ MTL). The input of MTL, immediately goes through one/multiple\footnote{Multiple FMS have the same time-domain receptive field, which is equal to the length of the input. This is contrary to the implementations of the transformer model that split the sequence based on number of attention heads.} Factorized Multimodal Self-attentions (FMS). The operations inside a single Factorized Multimodal Self-attention is demonstrated in Figure \ref{fig:attention}. For $3$ modalities\footnote{Remarks on more than $3$ modalities is in Appendix \ref{sec:app_remarks_more_3}.}, there exist $7$ distinct attentions inside a single FMS unit. Each attention has a unique receptive field with respect to modalities $f \in F=\{\texttt{L,V,A,LV,LA,VA,LVA}\}$; essentially denoting the modalities visible to the attention. Using this factorization, FMS explicitly accounts for possible unimodal, bimodal and trimodal interactions existing within the multimodal input space. All attentions within a FMS extend to the length of the sequence, and therefore can extract asynchronous relations within and across modalities. For $f\in F$, each attention within a single FMS unit is controlled by the Key $K^f$, Query $Q^f$, and Value $V^f$ all with dimensionality $\mathbb{R}^{\mathcal{T} \times \mathcal{T}}$; parameterized respectively using affine maps $W_{K^f}$, $W_{Q^f}$, and $W_{V^f}$. After the attention is applied using Key, Query and Value operations~\citep{vaswani2017attention}, the output of each of the attentions goes through a residual addition with its perceived input (input in the attention receptive field), followed by a normalization. 

The output of the FMS contains the aligned and extracted information from the unimodal, bimodal and trimodal factors. This output is high-dimensional; essentially $\mathbb{R}^{4\times \mathcal{T} \times e_x}$ (each dimension within input of shape $\mathcal{T} \times e_x$ is present in $4$ factors). Our goal is to reduce this high-dimensional data using a mapping from $\mathbb{R}^{4 \times \mathcal{T} \times e_x} \mapsto \mathbb{R}^{\mathcal{T} \times e_x}$. Without overparameterizing the FMS, in practice, we observed this mapping can be efficiently done using a simple 1D convolutional network $\mathcal{S}1_{\texttt{M} \in \{\texttt{L},\texttt{V},\texttt{A}\}}(\cdot); \mathbb{R}^{4} \mapsto \mathbb{R}$. Internally, $\mathcal{S}1(\cdot)$ maps its input to multiple layers of higher dimensions and subsequently to $\mathbb{R}$. Using language as an example, $\mathcal{S}1_\texttt{L}$ moves across language modality dimensions $e_\texttt{L}$ for $t=1 \dots \mathcal{T}$ and summarizes the information across all the factors. The output of this summarization applied on all modality dimensions and timesteps, is the output of FMS, which has the dimensionality $\mathbb{R}^{\mathcal{T} \times e_x}$.

In practice, there can be various possible unimodal, bimodal or trimodal interactions within a multimodal input. For example, consider multiple sets of important interactions between $\texttt{L}$ and $\texttt{V}$ (e.g. smile + positive word, as well as eyebrows up + excited phrase), all of which need to be highlighted and extracted. A single FMS may not be able to highlight all these interactions without diluting its intrinsic attentions. Multiple FMS can be used inside a MTL to efficiently extract diverse multimodal interactions existing in the input data\footnote{We study the impact of number of FMS units inside MTL in Section \ref{sec:results}.}. Consider a total of $U$ FMS units inside a MTL. The output of each FMS goes through a feedforward network (for each timestamp $t$ of the FMS output). The output of this feedfoward network is residually added with its input, and subsequently normalized. The feedforward network is the same across all $U$ FMS units and timestamps $t$. Subsequently, the dimensionality of the output of the normalizations collectively is $\mathbb{R}^{U \times \mathcal{T} \times e_x}$. Similar to operations performed by $\mathcal{S}1$, a secondary summarization network $\mathcal{S}2_{\texttt{M} \in \{\texttt{L},\texttt{V},\texttt{A}\}}(\cdot); \mathbb{R}^{U} \mapsto \mathbb{R}$ can be used here. $\mathcal{S}2$ is also a 1D convolutional network that moves across modality dimensions and different timesteps to map $\mathbb{R}^{U \times \mathcal{T} \times e_x}$ to $\mathbb{R}^{\mathcal{T} \times e_x}$. The output of the secondary summarization network is the final output of MTL, and denoted as $\hat{x}_i^{k+1}$.

Let $\hat{x}_i^K=\langle \hat{l}^{K}_{(\cdot,i)},\hat{v}^{K}_{(\cdot,i)},\hat{a}^{K}_{(\cdot,i)}\rangle$ be the output of last MTL in the stack. For supervision, we feed this input one timestamp at a time as input to a Gated Recurrent Unit (GRU)~\citep{cho2014gru}. The prediction is conditioned on output at timestamp $t=\mathcal{T}$ of the GRU, using an affine map to $d_y$. 
\section{Experimental Methodology}

In this section, we discuss the experimental methodology including tasks, datasets, computational descriptors, and comparison baselines.

\begin{table}[t]
\begin{center}
\setlength{\tabcolsep}{7.5pt}
\renewcommand{\arraystretch}{1.1}
\begin{tabular}{l|cc|cc}
\hline 
\textit{Model} \textbackslash \textit{Metric} & \textit{BA} & \textit{F1} & \textit{MAE} & \textit{Corr} \\
\hline 
MV-LSTM~\citep{rajagopalan2016extending} &73.9/--   \ \ \ \ \   &74.0/--   \ \ \ \ \   & 1.019 & 0.601 \\
TFN~\citep{tensoremnlp17} & 73.9/-- \ \ \ \ \   & 73.4/--  \ \ \ \ \   & 1.040 & 0.633 \\
MARN~\citep{zadeh2018multi} & 77.1/--  \ \ \ \ \  & 77.0/--  \ \ \ \ \   & 0.968 & 0.625 \\
MFN~\citep{zadeh2018memory} & 77.4/--  \ \ \ \ \  & 77.3/--  \ \ \ \ \   & 0.965 & 0.632 \\
RMFN~\citep{liang2018multimodal} & 78.4/--  \ \ \ \ \  & 78.0/-- \ \ \ \ \    & 0.922 & 0.681 \\
RAVEN~\citep{wang2018words} & 78.0/--  \ \ \ \ \  & --/--    & 0.915 & 0.691 \\
MulT~\citep{tsai2019multimodal} &  \ \ \ \ \  --/83.0 &  \ \ \ \ \   --/82.8 & 0.87 & 0.698 \\
\hline
FMT (ours) &  \textbf{81.5/83.5} & \textbf{81.4/83.5} & \textbf{0.837} & \textbf{0.744} \\
\hline
\end{tabular}
\end{center}
\caption{\label{table:mosi_performance}FMT achieves superior performance over baseline models for CMU-MOSI dataset (multimodal sentiment analysis). We report BA (binary accuracy) and F1 (both higher is better), MAE  (Mean-absolute Error, lower is better), and Corr (Pearson Correlation Coefficient, higher is better). For BA and F1, we report two numbers: the number on the left side of ``/'' is  calculated based on approach taken by \cite{zadeh2018multi}, and the right side is by \cite{tsai2019multimodal}.}

\end{table}

\subsection{Tasks and Datasets}
\label{sec:datasets}

The following inherently multimodal tasks (and accompanied datasets) are studied in this paper. All the tasks are related to multimodal language: a complex and idiosyncratic sequential multimodal signal, where semantics are arbitrarily scattered across modalities~\citep{holler2019multimodal}. 
 
\textit{Multimodal Sentiment Analysis:} The first benchmark in our experiments is multimodal sentiment analysis, where the goal is to identify a speaker's sentiment based on the speaker's display of verbal and nonverbal behaviors. We use the well-studied CMU-MOSI (CMU Multimodal Opinion Sentiment Intensity) dataset for this purpose~\citep{zadeh2016mosi}. There are a total of 2199 data points (opinion utterances) within CMU-MOSI dataset. The dataset has real-valued sentiment intensity annotations in the range $[-3,+3]$. It is considered a challenging dataset due to speaker diversity (1 video per distinct speaker), topic variations and low-resource setup.

\textit{Multimodal Emotion Recognition:} The second benchmark in our experiments is multimodal emotion recognition, where the goal is to identify a speaker's emotions based on the speaker's verbal and nonverbal behaviors. We use the well-studied {IEMOCAP} dataset~\citep{Busso2008IEMOCAP:Interactiveemotionaldyadic}. IEMOCAP consists of 151 sessions of recorded dialogues, of which there are 2 speaker's per session for a total of 302 videos across the dataset. We perform experiments for discrete emotions~\citep{ekman1992argument} of Happy,  Sad, Angry and Neutral (no emotions) - similar to previous works~\citep{tsai2019multimodal,wang2018words}.

\textit{Multimodal Speaker Traits Recognition:} 
The third benchmark in our experiments is speaker trait recognition based on communicative behavior of a speaker. It is a particularly difficult task, with 16 different speaker traits in total. We study the POM dataset which contains 1,000 movie review videos~\citep{Park:2014:CAP:2663204.2663260}. Each video is annotated for various personality and speaker traits, specifically: Confident (Con), Passionate (Pas), Voice Pleasant (Voi), Dominant (Dom), Credible (Cre), Vivid (Viv), Expertise (Exp), Entertaining (Ent), Reserved (Res), Trusting (Tru), Relaxed (Rel), Outgoing (Out), Thorough (Tho), Nervous (Ner), Persuasive (Per) and Humorous (Hum). The short form of these speaker traits is indicated inside the parentheses and used for the rest of this paper. 

\begin{table}[t]
\begin{center}
\setlength\tabcolsep{5.9pt}
\renewcommand{\arraystretch}{1.1}
\begin{tabular}{l|c|c|c|c|c|c|c|c}
\hline
\textit{Model} \textbackslash \ \textit{Emotion} & \multicolumn{2}{c}{\textit{Happy}} & \multicolumn{2}{|c}{\textit{Sad}} & \multicolumn{2}{|c}{\textit{Angry}} & \multicolumn{2}{|c}{\textit{Neutral}} \\
\hline

\textit{Metric}   & \textit{BA} & \textit{F1} & \textit{BA} & \textit{F1} & \textit{BA} & \textit{F1} & \textit{BA} & \textit{F1} \\ 
\hline
MV-LSTM~\citep{rajagopalan2016extending}   		& 85.9 & 81.3 & 80.4 & 74.0 & 85.1 & 84.3 & 67.0 & 66.7 \\
MARN~\citep{zadeh2018memory}			& 86.7 & 83.6 & 82.0 & 81.2 & 84.6 & 84.2 & 66.8 & 65.9 \\
MFN~\citep{zadeh2018memory}				& 86.5 & 84.0 & 83.5 & 82.1 & 85.0 & 83.7 & 69.6 & 69.2 \\
RMFN~\citep{liang2018multimodal} & 87.5 & 85.8 & 82.9 & 85.1 & 84.6 & 84.2 & 69.5 & 69.1 \\
RAVEN~\citep{wang2018words} & 87.3 & 85.8 & 83.4 & 83.1 & 87.3 & 86.7 & 69.7 & 69.3 \\ 
MulT~\citep{tsai2019multimodal} & \textbf{90.7} & \textbf{88.6} & 86.7 & 86.0 & 87.4 & 87.0 & 72.4 & 70.7 \\ 
\hline
FMT  & 88.8 & 87.2 & \textbf{88.0} & \textbf{87.7} & \textbf{89.7} & \textbf{89.5} & \textbf{74.0} & \textbf{73.8} \\  
\hline
\end{tabular}
\caption{FMT achieves superior performance over baseline models (with the exception of Happy emotion) for discrete emotions in IEMOCAP (multimodal emotion recognition). We report BA (binary accuracy) and F1 (both higher is better).}
\label{table:iemocap}
\end{center}
\end{table}

\begin{table*}[t]
\centering
\setlength\tabcolsep{4.8pt}
\renewcommand{\arraystretch}{1.1}
\begin{tabular}{l | c|c|c|c|c|c|c|c}
\hline
\multirow{2}{*}{\textit{Model} \textbackslash \ \textit{Trait}} & \textit{Con} & \textit{Pas}& \textit{Voi}& \textit{Dom}& \textit{Cre}& \textit{Viv}& \textit{Exp}& \textit{Ent}\\
\cline{2-9} 
& \textit{MA7} & \textit{MA7}& \textit{MA7}& \textit{MA7}& \textit{MA7}& \textit{MA7}& \textit{MA7}& \textit{MA7}\\
\hline
MV-LSTM~\citep{rajagopalan2016extending} & 25.6 & 28.6 & 28.1 & 34.5 & 25.6 & 32.5 & 32.5 & 29.6  \\
TFN~\citep{tensoremnlp17} & 24.1 & 31.0 & 31.5 & 34.5 & 24.6 & 25.6 & 27.6 & 29.1  \\
MARN~\citep{zadeh2018multi} & 29.1 & 33.0 & - & - & 31.5 & - & - & -  \\
MFN~\citep{zadeh2018memory} & 34.5 & 35.5 & 37.4 & 41.9 & 34.5 & 36.9 & 36.0 & 37.9  \\
RMFN~\citep{liang2018multimodal} & 37.4 & 38.4 & 37.4 & - & 37.4 & 38.9 & 38.9 & -  \\
MulT~\citep{tsai2019multimodal} & 34.5 & 34.5 & 36.5 & 38.9 & 37.4 & 36.9 & 37.9 & 39.4 \\
\hline
FMT & \textbf{40.9} & \textbf{42.4} & \textbf{42.4} & \textbf{44.3} & \textbf{41.4} & \textbf{39.4} & \textbf{41.4} & \textbf{39.4}  \\ 
\hline
\end{tabular}
\begin{tabular}{l | c|c|c|c|c|c|c|c}
\hline
\multirow{2}{*}{\textit{Model} \textbackslash \ \textit{Trait}} & \textit{Res} & \textit{Tru}& \textit{Rel}&\textit{Out}& \textit{Tho}& \textit{Ner}& \textit{Per}&\textit{ Hum}\\
\cline{2-9} 
& \textit{MA5} & \textit{MA5}& \textit{MA5}& \textit{MA5}& \textit{MA5}& \textit{MA5}& \textit{MA7}& \textit{MA5}\\
\hline
MV-LSTM~\citep{rajagopalan2016extending} & 33.0 & 52.2 & 50.7 & 38.4 & 37.9 & 42.4 & 26.1 & 38.9  \\
TFN~\citep{tensoremnlp17} & 30.5 & 38.9 & 35.5 & 37.4 & 33.0 & 42.4 & 27.6 & 33.0  \\ 
MARN~\citep{zadeh2018multi} & 36.9 & - & 52.2 & - & - & 47.3 & 31.0 & 44.8  \\
MFN~\citep{zadeh2018memory} & 38.4 & 57.1 & 53.2 & 46.8 & 47.3 & 47.8 & 34.0 & 47.3  \\ 
RMFN~\citep{liang2018multimodal} & 39.4 & - & 53.7 & - & 48.3 & 48.3 & 35.0 & 46.8  \\
MulT~\citep{tsai2019multimodal}& 41.4 & 60.6 & 54.2 & 43.3 & 49.3 & 46.3 & 33.5 & 43.3\\
\hline
FMT & \textbf{44.8} & \textbf{61.1} & \textbf{57.6} & \textbf{51.7} & \textbf{51.7} & \textbf{51.2} & \textbf{40.4} & \textbf{48.3}  \\ 
\hline
\end{tabular}
\caption{FMT achieves superior performance over baseline models in POM dataset (multimodal personality traits recognition). For label abbreviations please refer to Section \ref{sec:baseline_models}. MA(5,7) denotes multi-class accuracy for (5,7)-class personality labels (higher is better).  }
\label{table:pom}
\end{table*}

\subsection{Multimodal Computational Descriptors}
\label{sec:compdisc}

The following computational descriptors are used by FMT and baselines (all the baselines use the same descriptors in their original respective papers). 

\noindent \textit{Language:} P2FA forced alignment model~\citep{P2FA} is used to align the text and audio at word level. From the forced alignment, the timing of words and sentences are extracted. Word-level alignment is used to unify the modality frequencies~\citep{chen2017multimodal}. GloVe embeddings~\citep{pennington2014glove} are subsequently used for word representation. 

\noindent \textit{Visual:} For the visual modality, the Emotient FACET~\citep{emotient} is used to extract a set of visual features including Facial Action Units~\citep{ekman1980facial}, visual indicators of emotions, and sparse facial landmarks. 

\noindent \textit{Acoustic:} 
COVAREP~\citep{degottex2014covarep} is used to extract the following features: fundamental frequency, quasi open quotient~\citep{kane2013wavelet}, normalized amplitude quotient, glottal source parameters (H1H2, Rd, Rd conf)~\citep{drugman2012detection}, Voiced/Unvoiced segmenting features (VUV)~\citep{drugman2011joint}, maxima dispersion quotient (MDQ), the first 3 formants, parabolic spectral parameter (PSP), harmonic model and phase distortion mean (HMPDM 0-24) and deviations (HMPDD 0-12), spectral tilt/slope of wavelet responses (peak/slope), Mel Cepstral Coefficients (MCEP 0-24). 

\subsection{Baseline Models and Performance Measures}
\label{sec:baseline_models}

The following strong baselines are compared to FMT: \textit{MV-LSTM (Multi-view LSTM, \cite{rajagopalan2016extending})}, \textit{TFN (Tensor Fusion Network, \cite{tensoremnlp17})}, \textit{MARN (Multi-attention Recurrent Network, \cite{zadeh2018multi})}, \textit{MFN (Memory Fusion Network, \cite{zadeh2018memory})}, \textit{RAVEN (Recurrent Attended Variation Embedding Network, \cite{wang2018words})}, \textit{MulT\footnote{We use the aligned variant of MulT model, which has shown better performance than unaligned version in the original paper.} (Multimodal Transformer for [Un]aligned Sequences, \cite{tsai2019multimodal})}. There are fundamental distinctions between FMT and MulT, chief among them: 1) MulT consists of 6 transformers, 3 cross-modal transformers and 3 unimodal. Naturally this increases the overall model size substantially. FMT consists of only one transformer, with components to avoid overparameterization. 2) FMT sees interactions as undirected (unlike MulT which has $\texttt{L} \rightarrow \texttt{V}$ and $\texttt{V} \rightarrow \texttt{L}$), and therefore semantically combines two attentions in one. 3) MulT has no trimodal factors (which are important according to Section \ref{sec:results}). 4) MulT has no direct unimodal path (e.g. only $\texttt{L}$), as input to unimodal transformers are outputs of cross-modal transformers. 5) All FMT attentions have full time-domain receptive field, while MulT splits the input based on the heads. 

In their original publication, all the models report\footnote{With the exception of MulT for POM dataset, which is not reported in original paper. It is trained in this paper using authors' provided github code with hyperparameter search in Appendix \ref{sec:app_hyper}. } the performance over the datasets in Section~\ref{sec:datasets}, using the same descriptors discussed in Section~\ref{sec:compdisc}. The models in this paper are compared using the following performance measures (depending on the dataset): (BA) denotes binary accuracy - higher is better, (MA5,MA7) are 5 and 7 multiclass  accuracy - higher is better, (F1) denotes F1 score - higher is better, (MAE) denotes the Mean-Absolute Error - lower is better, (Corr) is Pearson Correlation Coefficient - higher is better. The hyperparameter space search for FMT (and baselines if retrained) is discussed in Appendix \ref{sec:app_hyper}.

\section{Results and Discussion}
\label{sec:results}

The results of sentiment analysis experiments on CMU-MOSI dataset are presented in Table~\ref{table:mosi_performance}. FMT achieves superior performance than the previously proposed models for multimodal sentiment analysis. We use two approaches for calculating BA and F1 based on negative vs. non-negative sentiment~\citep{zadeh2018multi} on the left side of /, and negative vs. positive~\citep{tsai2019multimodal} on the right side. MAE and Corr are also reported. For multimodal emotion recognition, experiments on IEMOCAP are reported in Table~\ref{table:iemocap}. The performance of FMT is superior than other baselines for multimodal emotion recognition (with the exception of Happy emotion). The results of experiments for personality traits recognition on POM dataset are reported in Table~\ref{table:pom}. We report MA5 and MA7, depending on the label. FMT outperforms baselines across all personality traits. 

We study the importance of the factorization in FMT. We first remove the unimodal, bimodal and trimodal attentions from the FMT model, resulting in 3 alternative implementations of FMT. Table \ref{table:mosi_mods} demonstrates the results of this ablation experiment over CMU-MOSI dataset. Furthermore, we use only one modality as input for FMT, to understand the importance of each modality (all other factors removed). We also replace the summarization networks with simple vector addition operation. All factors, modalities, and summarization components are needed for achieving best performance. 

We also perform experiments to understand the effect of number of FMT units within each MTL. Table \ref{table:mosi_heads} shows the performance trend for different number of FMT units. The model with $6$ number of FMS ($42$ attentions in total) achieves the highest performance ($6$ is also the highest number we experimented with). \cite{tsai2019multimodal} reports the best performance for CMU-MOSI dataset is achieved when using $40$ attentions per cross-modal transformer (3 of each, therefore $120$ attention, without counting the subsequent unimodal transformers). FMT uses fewer number of attentions than MulT, yet achieves better performance. We also experiment with number of heads for original transformer model~\citep{vaswani2017attention} and compare to FMT (Appendix \ref{sec:app_num_heads_baselines}).  

\section{Conclusion}

In this paper, we presented the Factorized Multimodal Transformer (FMT) model for multimodal sequential learning. Using a Factorized Multimodal Self-attention (FMS) within each Multimodal Transformer Layer (MTL), FMT is able to model the intra-model and inter-modal dynamics within asynchronous multimodal sequences. We compared the performance of FMT to baselines approaches over 3 publicly available datasets for multimodal sentiment analysis (CMU-MOSI, 1 label), emotion recognition (IEMOCAP, 4 labels) and personality traits recognition (POM, 16 labels). Overall, FMT achieved superior performance than previously proposed models across the studied datasets.

\begin{table}[t]
\begin{center}
\setlength{\tabcolsep}{7.5pt}
\renewcommand{\arraystretch}{1.2}
\begin{tabular}{l|cc|cc}
\hline 
\textit{Model} \textbackslash \textit{Metric} & \textit{BA} & \textit{F1} & \textit{MAE} & \textit{Corr} \\
\hline 
FMT [UNI] &   80.6/82.8 & 80.5/82.8 & 0.868 & 0.719 \\
FMT [BI] &  81.2/81.7 & 81.2/81.6 & 0.877 & 0.706 \\
FMT [TRI] &  80.2/81.6 & 80.2/81.5 & 0.874 & 0.705 \\
FMT [\texttt{L}] & 77.7/79.6 & 77.7/79.6 & 0.935 & 0.666 \\
FMT [\texttt{A}] & 62.5/62.7 & 62.6/73.2 & 1.338 & 0.306 \\
FMT [\texttt{V}] & 59.3/59.3 & 59.4/72.7 & 1.357 & 0.218 \\
FMT [$\mathcal{S}$] & 80.3/82.0 & 80.3/81.9 & 0.860 & 0.734 \\
\hline
FMT &  \textbf{81.5/83.5} & \textbf{81.4/83.5} & \textbf{0.837} & \textbf{0.744} \\
\hline
\end{tabular}
\end{center}
\caption{\label{table:mosi_mods} FMT ablation studies on CMU-MOSI dataset. UNI, BI, TRI denote removing all unimodal, bimodal and trimodal factors respectively. \texttt{L}, \texttt{A}, \texttt{V} denote using only language, audio, and visual factors respectively. $\mathcal{S}$ denotes the model with summarization networks replaced by simple addition. All factors, modalities, and components are needed for achieving best performance. }
\end{table}
\begin{table}[t]
\begin{center}
\setlength{\tabcolsep}{7.5pt}
\renewcommand{\arraystretch}{1.2}
\begin{tabular}{l|cc|cc}
\hline 
\textit{Model} \textbackslash \textit{Metric} & \textit{BA} & \textit{F1} & \textit{MAE} & \textit{Corr} \\
\hline 
FMT [1] &   80.0/82.0 & 79.4/81.1 &  0.864 &  0.712 \\
FMT [2] &  79.7/82.2 &  79.7/82.2 &  0.863 &  0.725 \\
FMT [3] &  79.2/80.9 &  79.1/80.8 &  0.905 &  0.698 \\
FMT [4] &  79.4/81.1 &  79.4/81.0 &  0.855 &  0.733 \\
FMT [5] &  81.5/82.6 &  81.5/82.5 &  0.886 &  0.711 \\
\hline
FMT [6] &  \textbf{81.5/83.5} & \textbf{81.4/83.5} & \textbf{0.837} & \textbf{0.744} \\
\hline
\end{tabular}
\end{center}
\caption{\label{table:mosi_heads} Multimodal sentiment analysis experiments with different number of FMS units inside MTL. The number in the square bracket indicates the number of FMS. The total number of attention is $7$ times the number in bracket.}

\end{table}

\clearpage

{\small
\bibliography{iclr2020_conference}
\bibliographystyle{iclr2020_conference}
}

\clearpage

\appendix

\section{Appendix}

\subsection{Training Remarks and Hyperparameter Space Search}\label{sec:app_hyper}
The hyperparameters of FMT include the Adam~\citep{kingma2014adam} learning rate ($\{0.001,0.0001\}$), structure of summarization network (randomly picked 5 architectures from $\{1,2,3\}$ layers of conv, with kernel shapes of $\{2,5,10,15,20\}$), number of MTL layers ($\{4,6,8\}$ except for ablation experiments which was $2\dots8$), number of FMT units ($\{4,6\}$, except for ablation experiment which was $1\dots6$), $e_{M \in \{\texttt{L},\texttt{V},\texttt{A}\}}$ ($\{20,40\}$), dropout ($0, 0.1$). The same parameters (when applicable) are used for training MulT for POM dataset (e.g. num encoder layers same as number of MTL). Furthermore, for MulT specific hyperparameters, we use similar values as Table 5 in the original paper. All models are trained for a maximum of $200$ epochs. The hyperparameter validation is similar to ~\cite{zadeh2018multi}.
\begin{table}[b]
\begin{center}
\setlength{\tabcolsep}{7.5pt}
\renewcommand{\arraystretch}{1.2}
\begin{tabular}{l|cc}
\hline 
\textit{Model} \textbackslash \textit{Metric} & \textit{MAE} & \textit{Corr}  \\
\hline 
FMT [2] & 0.881 & 0.720 \\
FMT [3] & 0.876 & 0.727 \\
FMT [4] & 0.871 & 0.723 \\
FMT [5] & 0.876 & 0.724 \\
FMT [6] & 0.852 & 0.730 \\
FMT [7] & 0.859 & 0.732 \\
\hline
FMT [8] & \textbf{0.837} & \textbf{0.744} \\
\hline
\end{tabular}
\end{center}
\caption{\label{table:mosi_layers} Multimodal sentiment analysis experiments with different number of MTL layers. The number in the square bracket indicates the number of MTL layers.}

\end{table}
\subsection{Number of MTL}
\label{sec:app_num_mtl}
We study the effect of number of MTL on FMT performance. Table \ref{table:mosi_layers} shows the results of this experiment. The best performance is achieved using $8$ MTL layers (which was also the maximum layers we tried in our hyperparameter search). 

\subsection{Number of Attention Heads for Original Transformer Model}
\label{sec:app_num_heads_baselines}

In this section, we discuss the effect of increasing the number of heads on the original transformer model (OTF, \cite{vaswani2017attention}). Please note that we implement the OTF to allow for all attention heads to have full input receptive field (from $1\dots\mathcal{T}$), similar to FMT. We increase the attention heads from $1$ to $35$ (after $35$ does not fit on a Tesla-V100 GPU with batchsize of $20$). Table \ref{table:baseline_heads} shows the results of increasing number of attention heads for both models. We observe that achieving superior performance is not a matter of increasing the attention heads. Even using $1$ FMS unit, which leads to $7$ total attention, FMT achieves higher performance than counterpart OTF.

\begin{table}[t!]
\begin{center}
\setlength{\tabcolsep}{7.5pt}
\renewcommand{\arraystretch}{1.2}
\begin{tabular}{l|cc|cc}
\hline 
\textit{Model} \textbackslash \textit{Metric} & \textit{BA} & \textit{F1} & \textit{MAE} & \textit{Corr} \\
\hline 
OTF [1] & 74.6/76.5 & 74.5/76.5 & 0.983 & 0.651 \\
OTF [2] & 76.8/78.8 & 76.6/78.9 & 0.975 & 0.655 \\
OTF [3] & 74.2/75.8 & 74.0/75.9 & 0.998 & 0.647 \\
OTF [4] & 76.5/77.9 & 76.6/78.2 & 1.022 & 0.632 \\
OTF [5] & 75.1/76.7 & 75.1/76.6 & 1.026 & 0.626 \\
OTF [6] & 71.6/72.3 & 71.3/72.6 & 1.094 & 0.677 \\
OTF [7] & 77.4/79.1 & 77.4/79.0 & 0.988 & 0.646 \\
OTF [14] & 77.0/78.5 & 76.9/78.5 & 0.972 & 0.683 \\
OTF [21] & 75.9/77.7 & 75.8/77.9 & 0.930 & 0.682 \\
OTF [35] & 67.6/68.9 & 67.2/73.2 & 1.174 & 0.502 \\
\hline
\end{tabular}
\end{center}
\caption{\label{table:baseline_heads} Results of experiments with different number of heads for OTF. The number in the square bracket indicates the number of heads.}

\end{table}

\subsection{Training Remarks for More than 3 Modalities}
\label{sec:app_remarks_more_3}

In many scenarios in nature, as well as what is currently pursued in machine learning, the number of modalities goes as high as 3 (mostly language, vision and acoustic, as studied in this paper). This leads to 7 attentions within each FMS, well manageable for successful training of FMT as demonstrated in this paper. However, as the number of modalities increases, the underlying multimodal phenomena becomes more challenging to model. This causes complexities for any competitive multimodal model, regardless of their internal design. While studying these cases are beyond the scope of this paper, due to rare nature of having more than 3 main modalities modalities, for FMT, the complexity can be managed due to the factorization in FMS. We propose two approaches: 1) for high number of modalities, the involved factors can be reduced based on domain knowledge, the nature of the problem, and the assumed dependencies between modalities (e.g. removing factors between modalities that are deemed weakly related). Alternatively, without making assumptions about inter-modality dependencies, a greedy approach may be taken for adding factors; an approach similar to stepwise regression~\citep{kleinbaum1988applied}, iteratively adding the next most important factor. Using these two methods, the model can cope with higher number of modalities with a controllable compromise between performance and overparameterization. 

\end{document}